\documentclass{article}

    \usepackage[preprint]{neurips_2025}

\usepackage[utf8]{inputenc} % allow utf-8 input
\usepackage[T1]{fontenc}    % use 8-bit T1 fonts
\usepackage{hyperref}       % hyperlinks
\usepackage{url}            % simple URL typesetting
\usepackage{booktabs}       % professional-quality tables
\usepackage{amsfonts}       % blackboard math symbols
\usepackage{nicefrac}       % compact symbols for 1/2, etc.
\usepackage{microtype}      % microtypography
\usepackage{xcolor,xspace}         % colors
\usepackage{amssymb,amsmath,amsthm,comment}
\usepackage{graphicx}
\usepackage{amsmath,amsfonts,bm}

\def\eqref#1{equation~\ref{#1}}
\def\1{\bm{1}}

\DeclareMathAlphabet{\mathsfit}{\encodingdefault}{\sfdefault}{m}{sl}
\SetMathAlphabet{\mathsfit}{bold}{\encodingdefault}{\sfdefault}{bx}{n}

\newtheorem{theorem}{Theorem}

\newtheorem{assumption}{Assumption}
\newtheorem*{remark}{Remark}

\newcommand{\hx}[1]{{\color{red} #1}}

\newcommand{\ltoo}{learn-to-optimize\xspace} 
\newcommand{\cwss}{CWSS\xspace} 

\title{A Learn-to-Optimize Approach for Coordinate-Wise Step Sizes for Quasi-Newton Methods}

\author{Wei Lin\\
CSE, CUHK\\
{\tt\small louislin@link.cuhk.edu.hk}
\And
Qingyu Song\\
CSE, CUHK\\
{\tt\small qysong21@cse.cuhk.edu.hk}
\And  
Hong Xu\\
CSE, CUHK\\
{\tt\small hongxu@cse.cuhk.edu.hk}
}

\begin{document}

\maketitle

%!TEX root = neurips_2025.tex

\begin{abstract}

Tuning step sizes is crucial for the stability and efficiency of optimization algorithms. While adaptive coordinate-wise step sizes have been shown to outperform scalar step size in first-order methods, their use in second-order methods is still under-explored and more challenging. Current approaches, including hypergradient descent and cutting plane methods, offer limited improvements or encounter difficulties in second-order contexts. To address these limitations, we first conduct a theoretical analysis within the Broyden-Fletcher-Goldfarb-Shanno (BFGS) framework, a prominent quasi-Newton method, and derive sufficient conditions for coordinate-wise step sizes that ensure convergence and stability.
Building on this theoretical foundation, we introduce a novel learn-to-optimize (L2O) method that employs LSTM-based networks to learn optimal step sizes by leveraging insights from past optimization trajectories, while inherently respecting the derived theoretical guarantees.
Extensive experiments demonstrate that our approach achieves substantial improvements over scalar step size methods
and hypergradient descent-based method, offering up to 4$\times$ faster convergence across diverse optimization tasks.

\end{abstract} 
%!TEX root = neurips_2025.tex

\section{Introduction}
\label{sec:introduction}

% \qy{QY:

% 1. Paper Writing Flow:

% Our practice: Try to learn Hessian, learn BFGS, learn preconditioning matrix (current approach).

% In paper: You should make it seems non-trivial. For example:
% 1. Find a problem. 2. Weakness of existing approaches. 3. Your analysis. 4. Your methods.

% 2. Intro section
% \begin{enumerate}
%     \item TBD Background: Straightforward, gradient-based methods. step size.
%     \item Non learning approaches and weakness: 2, 2.3. Current problem: Hard to clearly differentiate motivation and your method (analysis). If pick analysis, the non-triviality should be stated in intro. If pick motivation, current statement should be refined.
%     \item Learning approaches (focus on second order): most are first order, no second order? Why are learning approaches better than the non-learning ones? (Transfer to your scenario: learning-based)
%     \item Weakness of learning-based approaches. (To emphasize your approach: Theoretical inspiration)
%     \item Your contributions: theorem and experiments.
% \end{enumerate}
% }

Step size is an essential hyperparameter in optimization algorithms. It determines the rate at which the optimization variables are updated, and greatly influences the convergence speed and stability of the optimization process. 
In \textit{first-order} gradient-based optimization, how to choose an appropriate step size is well studied: 
The step size is typically adjusted adaptively using past gradient information such as in AdaGrad \cite{duchi2011adaptive}, RMSProp \cite{hinton2019neural}, and Adam \cite{kingma2014adam} for stochastic optimization tasks. 
% Despite differences in details, these adaptive methods assign a single scalar step size to all optimization variables (e.g. model parameters in ML) which works well in practice. 
These methods have demonstrated significant efficacy across a range of machine learning applications by dynamically tailoring the update scale for each iteration.

% However, in high-dimensional optimization problems, individual parameters may have different sensitivities to the step size, and may benefit from tailored step sizes.
\begin{comment}
Optimization is fundamental to machine learning with great impact on model accuracy, training speed, and computational efficiency. Gradient-based \textit{first-order} methods are widely used in continuous optimization due to simplicity and effectiveness. 
Their performance depends heavily on selecting an appropriate step size, i.e. learning rate \cite{wright2006numerical}. 
Step size determines how efficiently a method converges to the optimal solution. 
For this reason, adaptive \hx{step size?} methods, such as AdaGrad, RMSProp, and Adam, have become the de facto solution in many deep learning applications. 
They improve convergence by adjusting the learning rate using past gradient information and assigning coordinate-wise step sizes that adapt to parameter sensitivities \cite{mcmahan2010adaptive,duchi2011adaptive, kingma2014adam, hinton2019neural}. 
% Despite becoming standard in many deep learning applications, these methods are limited to first-order techniques, which do not utilize second-order curvature information. This lack of curvature insight can constrain their effectiveness, particularly in high-dimensional optimization landscapes.
\end{comment}

Step size in \textit{second-order} methods received much less attention thus far. 
Second-order methods leverage the curvature information to adjust both the search direction and step size, offering faster convergence in number of iterations, at the cost of high computational complexity in calculating the Hessian (or its approximation) \cite{wright2006numerical}.
%  address some of these limitations by incorporating curvature information, typically through the Hessian or its approximations, allowing for more precise adjustment of both the search direction and step size. 
A natural and common approach for step size tuning here is line search, which iteratively adjusts a \textit{scalar} step size along the descent direction until certain conditions, such as the Armijo condition, are met \cite{armijo1966minimization}. 
% This approach offers reliable convergence in smooth optimization problems and adapts well to unknown smoothness parameters. 
% Similar to first-order methods, line search also uses a scalar step size for all variables, which may limit its efficiency in high-dimensional spaces, where individual parameters could benefit from tailored step sizes.

In contrast to scalar step size, we study the more general \textit{coordinate-wise} step sizes (\cwss) in this work, which allow for individual variables to have different step sizes. 
\cwss are beneficial since different optimization variables may have different sensitivities to the step size; scalar step size is obviously a special case. 
They have also been shown to improve convergence in first-order methods \cite{amid2022step, kunstner2023searching, duchi2011adaptive}.

In this work, we explore the impact of \cwss in the context of second-order methods, which remains largely unexplored to our knowledge.
% \hx{cwss in second order methods has not yet been studied.} 
We choose the Broyden-Fletcher-Goldfarb-Shanno (BFGS) method \cite{broyden1965class}, one of the most widely used second-order optimization methods, as the backbone method. 
BFGS belongs to the quasi-Newton family of methods that iteratively update an approximation of the Hessian matrix using gradient information to reduce the complexity.

We start our study by demonstrating that existing solutions to tune \cwss in first-order methods do not work well in second-order contexts.
The first such approach is hypergradient descent \cite{maclaurin2015gradient,masse2015speed}, which iteratively tunes step sizes using their gradients at each BFGS step. 
We show empirically that it provides only marginal gains after the initial few steps of BFGS. 
Moreover, cutting-plane techniques, which expand backtracking line search into multiple dimensions, iteratively refine step sizes within feasible sets narrowed down by hypergradient-based incisions \cite{kunstner2023searching}. This method essentially offers an approximation of the Hessian in a first-order framework, thus complicating its direct application to second-order methods, in which Hessian approximation is handled by BFGS update, and the step sizes are adjusted to improve the Hessian approximation.
Further, the intricate curvature within the Hessian presents additional challenges in plane cutting.

Therefore, we explore the \ltoo (L2O) paradigm \cite{andrychowicz2016learning} in this work.
L2O replaces handcrafted rules with data-driven machine learning models that can adaptively learn efficient strategies, tailoring optimization processes to specific problem structures \cite{andrychowicz2016learning, lv2017learning}. 
L2O has shown promising results in first-order optimization by predicting the optimal step sizes dynamically based on the current optimization state \cite{liu2023towards, song2024towards}.

The application of L2O in quasi-Newton methods presents challenges. 
Whereas in first-order approaches, the step size primarily regulates the update magnitude, in second-order methods, it also affects the precision of Hessian approximations \cite{wright2006numerical}. 
This dual role adds complexities to step size tuning. 
% \hxq{in later analysis, we should back up this claim about dual-role to echo this point. pls add something if this is not there} 
% \wlc{Wei: Here I want to highlight the intricate and crucial role of step size tuning in quasi-Newton methods. Although this dual nature of the step size is evident, it is difficult to analytically determining how step size impacts the Hessian approximation. Consequently, our design is primarily driven by the idea that this dual role leads to a more complex influence of the step size on the iterates $x_k$. This complexity necessitates a significantly shorter unroll length in our method (specifically, 1) compared to first-order methods, which typically employ lengths of 20-100 in math-L2O. I will be elaborated this point in Section 4.}
Consequently, the unconstrained exploration inherent in conventional L2O makes convergence and stability harder to achieve within second-order L2O frameworks.

To address these challenges, we provide a theoretical analysis of coordinate-wise step sizes within the BFGS framework. 
We begin by outlining essential theoretical requirements for effective \cwss, aiming to ensure reliable optimization outcomes. These include achieving guaranteed convergence to a solution, maintaining stable progress towards the optimum, and preserving the strong convergence rates inherent to BFGS method. Guided by these foundational principles, we then derive a set of sufficient conditions for the \cwss matrix.
They effectively define a ``safe operating region'', steering the learning process away from potentially unstable or divergent behaviours for better efficiency. 
While meeting these sufficient conditions ensures desirable properties like convergence, they do not determine the optimal strategy for fastest progress. Our L2O approach is therefore designed to learn the most effective step-size selection strategy within this theoretically defined safe region, leveraging insights from past optimization trajectories to accelerate performance.
% \hxq{again too vague; give a bad impression to reviewers to feel that the work is thin and not solid; put more details and specifics here pls}

% We establish specific convergence criteria and derive conditions that effective step sizes must satisfy. 
% This analysis identifies key elements of the optimization state that influence step size selection and provides bounded conditions on step sizes that ensure convergence. Additionally, we derive that the coordinate-wise step sizes should asymptotically converge to an identity matrix to achieve super-linear convergence rates near the optimum. 

Specifically, we propose a customized L2O model, featuring a LSTM network, to generate \cwss for BFGS method. Motivated by theoretical analysis, our model takes optimization variables, gradients, and second-order search directions as input.
Distinct from many first-order L2O approaches that utilize longer unrolling horizons \cite{liu2023towards}, our model is trained with more frequent parameter updates to better capture the immediate effects of step size tuning in the sensitive quasi-Newton context. The training objective minimizes the expected objective value at the next iteration, augmented by a regularization term designed to ensure the learned step sizes adhere to our theoretical conditions for stability and efficient convergence.

We summarize our key contributions as follows:
\begin{enumerate}
    \item We are the first to investigate coordinate-wise step size tuning in the context of second-order optimization methods, specifically the BFGS algorithm. 
    % \item We derive conditions that effective coordinate-wise step sizes should satisfy, ensuring convergence and stability based on desired optimization properties.
    \item We establish theoretical foundation by deriving sufficient conditions for \cwss in the BFGS algorithm, ensuring convergence and stability and forming the principled basis for our L2O approach.
    \item We propose a new L2O method to generate \cwss for the BFGS algorithm, integrating both theoretical principles and adaptive learning to guide the optimization process.
    \item We empirically demonstrate the significant advantages of our method through extensive experiments on diverse optimization tasks, including classic optimization problems as well as a more challenging neural network training scenario. Our approach consistently achieves substantial speedups, delivering up to 4$\times$ faster convergence when compared to classic backtracking line search and hypergradient descent methods. Notably, the performance advantage of our method typically becomes more pronounced as the problem dimensionality increases, highlighting its strong scalability. Furthermore, our method exhibits improved stability, evidenced by lower variance in performance across multiple runs.
    % \hxq{add more details to highlight}
\end{enumerate}

%!TEX root = neurips_2025.tex

\section{Preliminaries}
\label{sec:related_work}

In this chapter, we introduce the basics of second-order optimization methods, with a focus on BFGS. We show how step size tuning critically affects both the convergence and the quality of Hessian approximations. 
Then we establish the key assumptions that will support our analysis of \cwss in BFGS framework.

% In optimization theory and practice, particularly for gradient-based methods, the step size (or learning rate) is a critical hyperparameter that significantly influences the algorithm's performance. Its importance stems from its direct impact on the update rule:

% \begin{equation*}
%     x_{k+1} = x_k - \alpha_k \nabla f(x_k)
% \end{equation*}

% where $x_k$ is the parameter vector at iteration $k$, $\alpha_k$ is the step size, and $\nabla f(x_k)$ is the gradient of the objective function $f$ at $x_k$.

% The step size directly affects the trade-off between convergence speed and stability.A larger step size can lead to faster convergence by taking bigger steps towards the optimum. For simple quadratic problems, the optimal fixed step size is $\alpha^* = 2 / (\lambda_{min} + \lambda_{max})$, where $\lambda_{min}$ and $\lambda_{max}$are the smallest and largest eigenvalues of the Hessian, respectively. However, if the step size is too large, it can cause overshooting and instability. The upper bound for stability in gradient descent is typically $\alpha < 2 / \lambda_{max}$.

% In practice, the optimal step size often varies during the optimization process. Early in training, larger step sizes can quickly move parameters to promising regions, while smaller step sizes are beneficial near the optimum for fine-tuning.

\subsection{Second-Order Methods}
\label{sec:second_order_methods}
Second-order optimization methods, such as Newton's method \cite{atkinson1991introduction}, utilize both gradient and curvature information to find the minimum of an objective function. While first-order methods typically achieve a sub-linear convergence rate \cite{beck2017first}, second-order methods generally exhibit a faster, quadratic convergence rate \cite{wright2006numerical}. 
% This significant difference makes second-order methods particularly advantageous in high-dimensional optimization problems, allowing for more efficient progress towards the optimum \cite{battiti1992first}.
In Newton's method, the objective function is locally approximated by a quadratic function around the current parameter vector $x_k$:
\begin{equation*}
    g(y) \approx f(x_k) + \nabla f(x_k)^T(y-x_k) + \frac{\alpha_k}{2} (y-x_k)^T H_k (y-x_k),
\end{equation*}
where $H_k$ is the Hessian matrix and $\alpha_k$ is the damped parameter. By minimizing the quadratic approximation, the update rule for Newton's method becomes \cite{wright2006numerical}:
\begin{equation*}
    x_{k+1} = x_k - \alpha_k H_k^{-1} \nabla f(x_k).
\end{equation*}

Computing the Hessian is quite expensive and often infeasible for large-scale problems \cite{pearlmutter1994fast}. 
Instead, quasi-Newton methods were proposed to approximate the Hessian to be more affordable and scalable \cite{dennis1977quasi,broyden1967quasi}. 
Generally, quasi-Newton methods maintain an approximation of the Hessian matrix $B_k\approx H_k$ at each iteration, updating it with a rank one or rank two term based on the gradient differences between two consecutive iterations \cite{conn1991convergence,broyden1965class}. During this process, the Hessian approximation is restricted to follow the secant equation \cite{wright2006numerical}:
\begin{equation}
\label{eq:secant condition}
    B_{k+1} s_k = y_k,
\end{equation}
where $s_k = x_{k+1} - x_k$ and $y_k = \nabla f(x_{k+1}) - \nabla f(x_k)$. In the most common BFGS method, the Hessian approximation $B_k$ is updated at each iteration using the formula:
\begin{equation*}
    B_{k+1} = B_k - \frac{B_k s_k s_k^T B_k} {s_k^T B_k s_k} + \frac{y_k y_k^T} {y_k^T s_k},
\end{equation*}
and the update rule becomes:
\begin{equation*}
    x_{k+1} = x_k - \alpha_k B_k^{-1} \nabla f(x_k).
\end{equation*}

Although the enrollment of curvature information can greatly assist the optimization process, it also makes the algorithm more sensitive to the step size selection \cite{wills2018stochastic}. The step size influences the update of the Hessian approximation, and an inappropriately large step can lead to violations of the curvature condition $y_k^T s_k > 0$, potentially resulting in an indefinite Hessian approximation \cite{wright2006numerical}. 
% The quality of the Hessian approximation is also heavily dependent on the step size. If $\alpha_k$ is too small, the changes in gradients $y_k$ and parameters $s_k$ may be seriously influenced by numerical errors, leading to poor Hessian updates and slower convergence \cite{dennis1984inaccuracy}. In quasi-Newton methods,
The step size must balance between exploiting the current curvature information (encoded in $B_k$) and allowing for sufficient exploration of the parameter space. This balance is more delicate than in first-order methods due to the adaptive nature of the search direction.

\subsection{Assumptions}

Our objective is to minimize the convex objective function $f(x)$ over $x\in \mathbb R^n$: $
    \min_{x \in \mathbb{R}^n} f(x).
$
Our analysis relies on the following standard assumptions regarding the objective function $f$ and the Hessian approximations $B_k$. These assumptions are common in optimization literature~\cite{song2024towards,liu2023towards,wright2006numerical}:

\begin{assumption}
    The objective function $f$ is $L$-smooth, meaning there exists a constant $L$ such that:
    \begin{equation*}
        \| \nabla f(x) - \nabla f(y) \| \leq L \| x - y \|.
    \end{equation*}
\end{assumption}

\begin{assumption}
    The gradient $\nabla f(x)$ is differentiable in an open, convex set $D$ in $\mathbb{R}^n$, and $\nabla^2 f(x)$ is continuous at the minimizer $x^*$ with $\nabla^2 f(x^*)$ being nonsingular.
\end{assumption}

\begin{assumption}
    The Hessian approximation generated by BFGS method is positive definite. Furthermore, there exists a constant $M\geq 1$ such that:
    \begin{equation*}
         \text{cond}(B_k)= \lambda_{\max}(B_k)/\lambda_{\min}(B_k) \leq M,
    \end{equation*}
    where $\lambda_{\min}(B_k)$ and $\lambda_{\max}(B_k)$ are the smallest and largest eigenvalues of $B_k$, respectively. By this assumption we assume the Hessian approximation remain well-conditioned. 
\end{assumption}

\begin{assumption}
\label{assumption:bounded_update_direction}
    The norm of update direction $B_k^{-1}\nabla f(x_k)$ is upper bounded by a constant $R$:
    \begin{equation*}
        \| B_k^{-1} \nabla f(x_k) \| \leq R.
    \end{equation*}
    This is a standard assumption in the analysis of quasi-Newton methods, as $B_k^{-1}$ is maintained bounded through stable Hessian approximations \cite{broyden1967quasi}, and gradients $\nabla f(x_k)$ typically diminish near optimal points, ensuring the update direction remains controlled. 
\end{assumption}

\section{Coordinate-Wise Step Sizes for BFGS}
\label{sec:coordinate_wise_step_size}

% \qy{
% Main parts of this section:
% \begin{enumerate}
%     \item Why choose vector step size: Observation (another name?), proof.
%     \item How should a good vector step size be like: Thereom 1, 2, and 3. State their insight for implementation.
% \end{enumerate}
% }

In this section, we first analyze the theoretical advantages of \cwss and then explore hypergradient descent as a practical method for its tuning. However, the limited improvements achieved through hypergradient descent reveal the challenges of finding effective \cwss, prompting us to consider alternative approaches. We resort to L2O method that can directly learn the step sizes from data derived from similar optimization problems. Building on this perspective, we establish sufficient conditions for effective \cwss that ensure convergence and descent properties, thus laying a solid foundation for learning-based approaches that can predict optimal step sizes efficiently during optimization. 

% This learning-based approach has the advantage of bypassing the need for additional evaluations of the objective function and gradients, providing a more efficient path to effective step size determination. Despite these benefits, a purely black-box L2O method lacks interpretability and can result in unpredictable behavior, limiting its applicability. Therefore, rather than adopting a black-box approach, we construct a more transparent, interpretable, and math-inspired L2O method. This method combines the flexibility of data-driven learning with the rigor of mathematical principles, offering a structured way to inform and guide optimization decisions.

% This integration of theory and data-driven adaptability not only enhances the interpretability and robustness of our method but also advances the L2O paradigm in a more mathematically grounded direction.

\subsection{Gain of Coordinate-Wise Step Sizes}
\label{sec:simple_example}

To illustrate the potential benefits of \cwss in the BFGS method, let us consider the theoretical implications of relaxing the constraint that step size should be a scalar. Assume we have identified an optimal scalar step size, denoted by $\alpha_k^*$, for the $k$-th iteration. 
% Since the restriction of a convex function to a line remains convex \cite{boyd2004convex}, this optimal step size $\alpha_k^*$ satisfies:
% \begin{equation}
% \label{eq:scalar_step_size}
% \begin{aligned}
% &\frac{d}{d \alpha_k} f(x_{k+1}) \bigg|_{\alpha_k = \alpha_k^*} \\
% =& \frac{d}{d\alpha_k^*} f(x_k - \alpha_k ^*B_k^{-1} \nabla f(x_k)) \\
% =& -\nabla f(x_k - \alpha_k^* B_k^{-1} \nabla f(x_k))^\top B_k^{-1} \nabla f(x_k) \\
% =& 0.
% \end{aligned}
% \end{equation}
% When constrained to a scalar form, $\alpha_k^*$ guarantees optimality along the single search direction $B_k^{-1} \nabla f(x_k)$. However, 
If we allow the step size to be a diagonal matrix $P_k$ rather than a scalar, the optimality condition of $\alpha_k^*$ may no longer hold. 
To explore this, we can set the coordinate-wise step sizes $P_k$ as:
\begin{equation*}
P_k = \alpha_k^* I - \frac{1}{LR} v_k B_k^{-1} \nabla f(x_k),
\end{equation*}
where $v_k = \text{diag}(\nabla f(x_k - \alpha_k^* B_k^{-1} \nabla f(x_k)))$, $L$ is the Lipschitz constant of $\nabla f$ and $R$ is from Assumption~\ref{assumption:bounded_update_direction}. This coordinate-wise step size $P_k$ is theoretically guaranteed to perform better than the scalar step size $\alpha_k^*$:
\begin{equation*}
\begin{split}
&f(x_k-P_kB_k^{-1}\nabla f(x_k)) \leq f(x_k-\alpha_k^*B_k^{-1}\nabla f(x_k)) \\
&\quad - \frac{1}{2LR} |\nabla f(x_k-\alpha_k^*B_k^{-1}\nabla f(x_k))\odot B_k^{-1}\nabla f(x_k) |^2.
\end{split}
\end{equation*}

This demonstrates that \cwss in the BFGS method can yield a more substantial decrease in the objective function than a scalar step size. A more detailed analysis is provided in Appendix~\ref{appendix:gain_of_cwss}.

\subsection{Numerical Analysis of Coordinate-Wise Step Size: A Hypergradient Descent Method}
\label{sec:hypergradient_descent}

\begin{table}[ht]
    \caption{Objective value of the least square problem with hypergradient descent (HGD) on $P_k$ for different BFGS iterations.}
    \label{tab:hypergradient_descent}
    \centering
    \begin{tabular}{ccccc}
        \toprule
        HGD (i)  &        1  &        5  &        10  &        20 \\
        BFGS (k) &           &           &           &           \\
        \midrule
        1                &  7.52938 &  6.32887 &  5.22274 &  4.32551 \\
        2                &  1.97834 &  1.95869 &  1.93509 &  1.89111 \\
        3                &  0.88499 &  0.88143 &  0.87703 &  0.86839 \\
        4                &  0.44807 &  0.44746 &  0.44670 &  0.44519 \\
        5                &  0.25669 &  0.25658 &  0.25644 &  0.25617 \\
        % 6                &  0.14995 &  0.14992 &  0.14988 &  0.14979 \\
        % 7                &  0.08901 &  0.08900 &  0.08899 &  0.08896 \\
        % 8                &  0.05425 &  0.05425 &  0.05424 &  0.05423 \\
        % 9                &  0.03373 &  0.03373 &  0.03372 &  0.03372 \\
        % 10               &  0.02107 &  0.02107 &  0.02107 &  0.02107 \\
        \bottomrule
        \end{tabular}   
        \vspace{-5mm}
\end{table}

% Coordinate-wise step sizes operate in a higher-dimensional search space compared to traditional line search methods, which are confined to a single dimension. Given $B_k^{-1}\nabla f(x_k)$, any vector in the space can be achieved as $P_kB_k^{-1}\nabla f(x_k)$ through an appropriate choice of $P_k$. There exists an optimal $P_k^*$ such that $P_k^* B_k^{-1}\nabla f(x_k) = x-x^*$, allowing the optimization to reach the solution in a single iteration. However, finding this optimal $P_k^*$ is as hard as finding $x^*$ itself. Therefore, we seek a computationally efficient approach to find the coordinate-wise step sizes $P_k$ that can enhance convergence.

Building upon section~\ref{sec:simple_example}, we investigate hypergradient descent on coordinate-wise step size matrix $P_k$. The update rule with \cwss takes the form:
\begin{equation}
\label{eq:bfgs_update}
x_{k+1} = x_k - P_k B_k^{-1}\nabla f(x_k).
\end{equation}
We initialize $P_k^{0}$ as the identity matrix $I$ and then perform hypergradient descent on $P_k$ using the gradient of $f(x_{k+1})$ with respect to $P_k^{i}$ to obtain $P_k^{i+1}$: 
\begin{equation*}
P_k^{i+1} = P_k^{i} - \eta \frac{\partial f(x_k - P_k^{i}B_k^{-1}\nabla f(x_k))}{\partial P_k^{i}},
\end{equation*}
where $\eta$ is the step size for the gradient descent on $P_k$.
After $T$ iterations, we employ $P_k^{T}$ in the update rule~\ref{eq:bfgs_update}. 

We conduct experiments on the least squares problem to assess the effectiveness of hypergradient descent applied to $P_k$. Each BFGS iteration includes 20 steps of hypergradient descent, after which the most recent $P_k$ identified by hypergradient descent is used in BFGS update. Table \ref{tab:hypergradient_descent} presents the experimental results, where each row shows the objective value within one BFGS iteration across different hypergradient descent steps. The results demonstrate that while hypergradient descent shows some improvement over standard BFGS, the benefits become increasingly marginal as iterations progress. This implies that finding an effective $P_k$ is inherently challenging. 
% From an alternating optimization perspective, convergence to $x^*$ can be achieved through updates to either $P_k^{i}$ or $x_k$. Hypergradient descent on $P_k^{i}$, despite being computationally lightweight as a first-order method, offers limited performance gains. In contrast, BFGS updates, while computationally more demanding due to their need to maintain accurate Hessian approximations, provide a more substantial optimization progress. 
% %\qy{"finding... " these two sentences seem unrelated to former and later. }
% It is reasonable to allocate the majority of computational resources to the BFGS process itself due to its fast convergence. 

This observation motivates exploring methods that can provide meaningful improvements without incurring significant computational costs. This leads us to consider a question: Can we leverage the patterns in optimization trajectories to generate effective step sizes directly? In many optimization scenarios, similar patterns of gradients and Hessian approximations may warrant similar step size adjustments. If these patterns could be learned from data, we might be able to bypass the iterative computation entirely. This insight motivates us to explore the L2O paradigm.

L2O has shown strong potential in capturing complex patterns and relationships, making it suitable for tasks like predicting step sizes based on optimization state features~\cite{liu2023towards}. By leveraging a neural network, L2O could potentially map the current optimization state to \cwss directly. This approach would allow immediate predictions of effective step sizes without iterative refinement. Before detailing our L2O model, we first establish theoretical conditions for \cwss in BFGS to ensure desirable properties like convergence and stability, which will guide our L2O design.

\subsection{Sufficient Conditions for Coordinate-Wise Step Sizes with Theoretical Guarantee}
\label{sec:conditions}

% Effective coordinate-wise step size must satisfy certain theoretical requirements to ensure that each iterative update is convergent. In this section, we establish the foundational conditions required for these coordinate-wise step sizes to achieve convergence properties. These conditions provide a systematic framework for the development of any adaptive coordinate-wise step size tuning methods that maintain the stability and efficiency of BFGS method.

Effective \cwss are crucial for ensuring that each BFGS iteration leads towards a solution. Rather than allowing the L2O model to determine these step sizes arbitrarily, which could lead to unpredictable behavior, we aim to unbox this process through theoretical guidance. This section lays the groundwork by identifying sufficient conditions that \cwss must satisfy for provable convergence and stability. By establishing these foundational principles, we provide a systematic basis for constraining and guiding L2O, ensuring adaptive step-size mechanisms enhance BFGS while preserving its desirable characteristics.

To ensure \cwss are theoretically sound and practically beneficial, we propose the following requirements:
\textit{\begin{enumerate}
    \item (Convergence Guarantee) The generated sequence ${x_k}$ converges to one of the local minimizers of $f$.
    \item (Stability Guarantee) Each update moves towards the minimizer.
    \item (Convergence Rate Guarantee) The method achieves superlinear convergence.
\end{enumerate}}
The first requirement, ensuring the generated sequence converges to a local minimizer, establishes a fundamental guarantee of reliable final outcomes, extending the concept of Fixed Point Encoding~\cite{ryu2022large}. The second requirement, instead, shifts focus to the optimization process itself, emphasizing directional accuracy to ensure stable progress by mandating that each update consistently moves towards the minimizer, thereby preventing detours or excessive zigzagging. Finally, the third requirement addresses convergence speed, aiming to preserve the characteristic superlinear convergence rate of the BFGS method \cite{wright2006numerical}, a key advantage we seek to maintain within our L2O framework.

We now present three theorems that provide sufficient conditions for coordinate-wise step sizes to satisfy the proposed requirements. The proofs are provided in Appendix~\ref{appendix:proofs}.

\begin{theorem}
\label{theorem:bounds of smooth function}
    Let $\{x_k\}$ be the sequence generated by \eqref{eq:bfgs_update}. If the coordinate-wise step size $P_k$ satisfies
    \begin{equation*}
    \begin{aligned}
        \|P_k\|_2 &\leq \frac{\alpha}{L\|B_k^{-1}\|_2}, \\
        \|P_k^{-1}\|_2 &\leq \frac{\|B_k^{-1}\nabla f(x_k)\|^2}{\beta\nabla f(x_k)^\top B_k^{-1} \nabla f(x_k)},
    \end{aligned}
    \end{equation*}
for certain $0<\alpha< 2$ and $\beta > 0$, where $L$ is the Lipschitz constant of gradients and $B_k$ is the approximate Hessian generated by BFGS, then the sequence of gradients converges to zero: $\lim _{k\to \infty}\|\nabla f(x_k)\|_2=0$.
\end{theorem}

\begin{remark}
Theorem \ref{theorem:bounds of smooth function} establishes sufficient conditions for gradient convergence while maintaining substantial implementation flexibility. The theorem's bounds on $P_k$ are particularly accommodating: the lower bound of its minimal eigenvalue is allowed to be close to zero through appropriate selection of $\beta$, while setting $\alpha$ near $2$ allows the upper bound of the maximal eigenvalue to approach $2/(L\|B_k^{-1}\|_2)$, which remains strictly less than 2. 
\end{remark}

Theorem~\ref{theorem:bounds of smooth function} suggests a pragmatic simplification: constraining the elements of the coordinate-wise step size to the interval between 0 and 2 should be sufficient for practical implementations. Moreover, theorem~\ref{theorem:bounds of smooth function} indicates that $P_k$ should be computed as a function of both the gradient $\nabla f(x_k)$ and Hessian approximation $B_k$, as evidenced by the presence of both gradient and Hessian information in bounds. Notably, these results extend beyond convex optimization, requiring only L-smoothness of the objective function rather than convexity.

\begin{theorem}
\label{theorem:bounds of convex function}
    Let $f:\mathbb R^n \to \mathbb R$ be a twice continuously differentiable convex function that has Lipschitz continuous gradient with $L>0$. Let $x^*$ denote the unique minimizer of $f$. Suppose that $\{B_k\}$ is a sequence of approximate Hessians such that they are uniformly lower bounded: $\gamma I \preceq B_k,$
    for certain constant $\gamma > 0$. Let $\{P_k\}$ be a sequence of diagonal matrices with entries $p_{k,i}$ satisfying:
    \begin{equation*}
        0 < p_{k,i} \leq \frac{2 \gamma}{L},
    \end{equation*}
    Define the iterative sequence $\{x_k\}$ by \eqref{eq:bfgs_update}. Then, the sequence $\{x_k\}$ satisfies:
    \begin{equation*}
        \|x_{k+1} - x^*\| \leq \|x_k - x^*\|.
    \end{equation*}
\end{theorem}

\begin{remark}
Since $B_k$ captures the average Hessian behavior between consecutive points $x_{k-1}$ and $x_k$, its eigenvalues lie within the bounds of $\nabla^2 f(x)$, yielding $ \gamma \leq L$. This relationship reveals that the seemingly restrictive upper bound $\frac{2\gamma}{L}$ for $p_{k,i}$ simplifies to 2. This aligns with Theorem \ref{theorem:bounds of smooth function}, as both theorems suggest the coordinate-wise step sizes should lie within the interval between 0 and 2. However, theorem~\ref{theorem:bounds of convex function} makes an additional assumption of convexity, which enables a stronger guarantee, i.e., each iteration strictly decreases the distance to the optimum.
\end{remark}

The theorem can be reduced to classical optimization methods in certain scenarios. For instance, setting $B_k = I$ and $P_k = \alpha I$ with $\alpha \leq 2/L$ yields the standard gradient descent method with constant step size. Moreover, the proof of theorem \ref{theorem:bounds of convex function} indicates that optimal $P_k$ values should minimize the spectral radius of $T_k = I - P_k B_k^{-1} H_k$. As the algorithm progresses ($k\to \infty$), $B_k$ approaches $H_k$, suggesting that $P_k$ should converge to the identity matrix. This convergence behavior is formally established in the subsequent theorem.

\begin{theorem}
\label{theorem:limit of P}
    Suppose $\nabla f(x)$ is differential in the open, convex set $D$ in $\mathbb R^n$, and $\nabla^2 f(x)$ is continuous at $x^*$ and $\nabla ^2f(x^*)$ is nonsingular. $\{B_k \}$ is generated by BFGS and $\{x_k\}$ is generated by the update rule. If $\{x_k\}$ converges to $x^*$. Then $\{x_k \}$ converges superlinearly to $x^*$ if  $\{ P_k\}$ converges to the identical matrix. 
\end{theorem}
% \begin{proof}
%     See Appendix \ref{proof:limit of P}.
% \end{proof}
Theorem \ref{theorem:limit of P} provides crucial insight into the asymptotic behavior of coordinate-wise step sizes. When the iterates are far from the optimum, coordinate-wise step sizes can accelerate convergence by adapting to the local geometry of the objective function. However, as the algorithm approaches the optimum, the BFGS method naturally provides increasingly accurate Hessian approximations. At this stage, additional coordinate-wise scaling becomes unnecessary and could potentially interfere with the superlinear convergence properties of BFGS. It suggests that adaptive schemes for $P_k$ should be designed to gradually reduce their influence as the optimization progresses, eventually allowing the natural BFGS updates to dominate near the optimum.

\section{L2O Model}
\label{sec:L2O_method}

% \qy{QY:
% \begin{enumerate}
%     \item Baseline: learning and non-learning (show the weakness of a learning without theoretical support)
%     \item Ablation: Demonstrate the alignment between theorem and experimental results.
% \end{enumerate}
% }

Building on the theoretical foundations established in the previous section, we now present our L2O model for \cwss tuning in BFGS optimization. Our design is guided by the derived theoretical conditions to ensure stability and convergence, while leveraging neural networks to adapt to the local optimization geometry. 

We propose an L2O method using an LSTM (Long Short-Term Memory) network to predict coordinate-wise step sizes \cite{liu2023towards}. The architecture is structured as follows:
\begin{align*}
    h_k, o_k &= \text{LSTM}(x_k, \nabla f(x_k), B_k^{-1}\nabla f(x_k), h_{k-1}, \phi_{\text{LSTM}}), \\
    p_k &= \text{MLP}(o_k, \phi_{\text{MLP}}), \\
    P_k &= \text{diag}(2\sigma(p_k)),
\end{align*}
where, $h_k$ is the LSTM hidden state, initialized randomly for the first iteration, and $o_k$ is the embedding output from the LSTM network. The parameters of the LSTM and MLP (Multi-Layer Perceptron) networks are denoted by $\phi_{\text{LSTM}}$ and $\phi_{\text{MLP}}$, respectively. 

% According to section~\ref{sec:conditions}, the values of $p_{k,i}$ should lie within the interval [0, 2]. Therefore, the activation function $\sigma$, which represents the sigmoid function, is employed to ensure that the predicted step sizes are within a suitable range, enabling bounded and adaptive adjustments for each coordinate.

A key aspect of our model's design is the enforcement of theoretically-informed bounds on the predicted step sizes. As established in Theorem~\ref{theorem:bounds of smooth function} and \ref{theorem:bounds of convex function}, specific bounds on $P_k$ are sufficient to guarantee convergence properties. Ideally, these theorems suggest bounds dependent on quantities like the Lipschitz constant $L$ or the Hessian conditioning $\gamma$. However, these parameters are often unknown or computationally prohibitive to estimate accurately during optimization. Consequently, as a practical and robust simplification suggested by the remarks , we constrain the elements of $P_k$ to lie within the interval between 0 and 2. This is achieved by using a scaled sigmoid activation function for the output. This hard constraint ensures that the predicted step sizes always reside within a safe range derived from our theoretical analysis. Within this bounded regime, the L2O model is then tasked with learning the more nuanced, data-driven strategy for selecting optimal \cwss that accelerate convergence while inherently respecting the stability conditions.

To enhance scalability and parameter efficiency, we employ a coordinate-wise LSTM approach, where the same network is shared across all input coordinates, as suggested in \cite{andrychowicz2016learning,lv2017learning}. This design allows the L2O method to adapt to problems of varying dimensionality without an increase in the number of parameters, making it highly efficient for large-scale optimization tasks.

The L2O model is trained on datasets of diverse optimization problems, allowing it to learn common structures and behaviors across different problem instances. The training process comprises two nested optimization loops that interact as follows:

\paragraph{Inner Optimization Loop}
In the inner loop, we utilize the current L2O model to optimize the given objective function. For each training instance, which is an optimization problem, we perform totally $K$ iterations. At each iteration $k$, the L2O model predicts the step size $P_k$, which is then used to update the current point. This process allows the L2O model to adaptively adjust the step size for each coordinate based on the current optimization state, effectively utilizing the curvature information encoded in $B_k^{-1}$.

\paragraph{Outer Optimization Loop}
Immediately after each iteration of the inner loop, we update the parameters of the neural networks ($\phi_{\text{LSTM}}$ and $\phi_{\text{MLP}}$) using backpropagation. The update is based on the objective function value achieved at $x_{k+1}$. While many first-order L2O methods use longer unrolling length before model updates \cite{andrychowicz2016learning,song2024towards,lv2017learning}, our approach updates more frequently. This is crucial because the step size $P_k$ in our quasi-Newton context not only directs the iterate $x_{k+1}$ but also critically shapes the Hessian approximation $B_{k+1}$. Such dual influence necessitates these frequent updates for the L2O model to effectively capture the immediate impact of its predictions on both the optimization trajectory and the Hessian quality.

The loss function used for training draws inspiration from the concept of exact line search, a technique in optimization that determines the optimal step size along a given search direction to minimize the objective function precisely. Specifically, we aim to minimize the objective function value in the next iteration with a regularization term:
\begin{equation*}
\min_{\phi_{\text{LSTM}}, \phi_{\text{MLP}}} \mathbb{E}_{f \sim \mathcal{F}} [f(x_{k+1})] + \lambda \| P_k - I \|^2_F
\end{equation*}
where $\mathcal{F}$ represents the distribution of optimization problems used for training and $\lambda$ is 
the regularization parameter. The regularization term ensures that as we approach the optimum, the coordinate-wise step sizes converge toward an identity matrix, aligning with the insight from Theorem~\ref{theorem:limit of P}. 

In BFGS method, the step size can be viewed as a correction to the Hessian approximation. In early optimization stages, the objective value primarily drives the loss function, and the Hessian approximation may lack precision. Thus, an adaptive \cwss is necessary to enhance the accuracy of the Hessian approximation based on the current state. However, as the optimization nears convergence, the Hessian approximation becomes more accurate, shifting the influence on the loss function to the regularization term. At this point, the \cwss converge to the identity matrix, as further corrections to the Hessian approximation are no longer required.

\section{Experiments}
\label{sec:experiments}

\begin{figure*}[tb]
    \vspace{-2mm}
    \centering
    \begin{minipage}{0.32\textwidth}
        \centering
        \includegraphics[width=\textwidth]{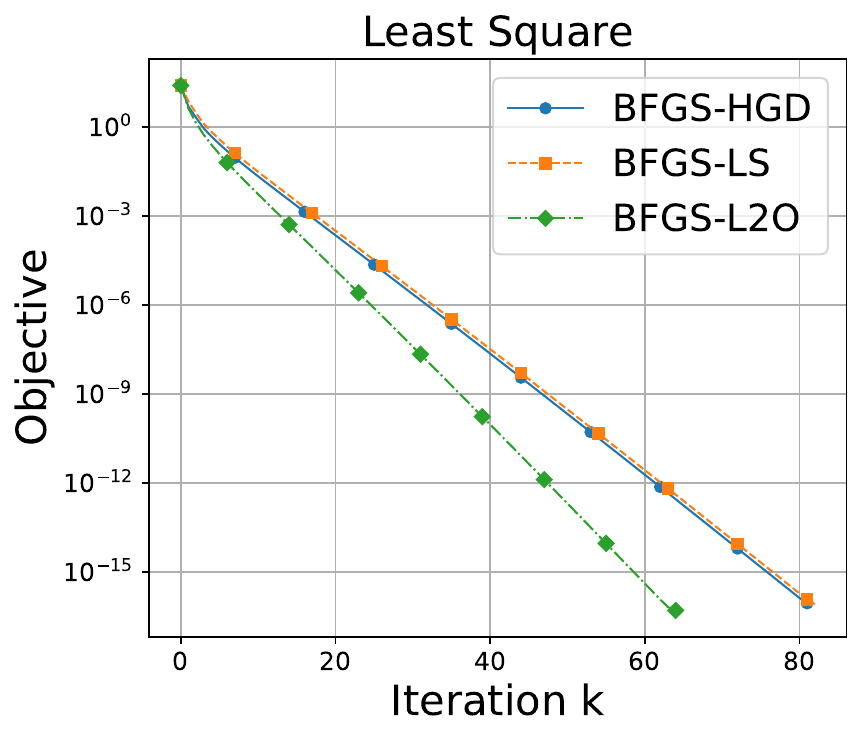}
        \vspace{-7mm}
        \caption{Least Squares.}
        \label{fig:least_square}
    \end{minipage}%
    \hfill
    \begin{minipage}{0.32\textwidth}
        \centering
        \includegraphics[width=\textwidth]{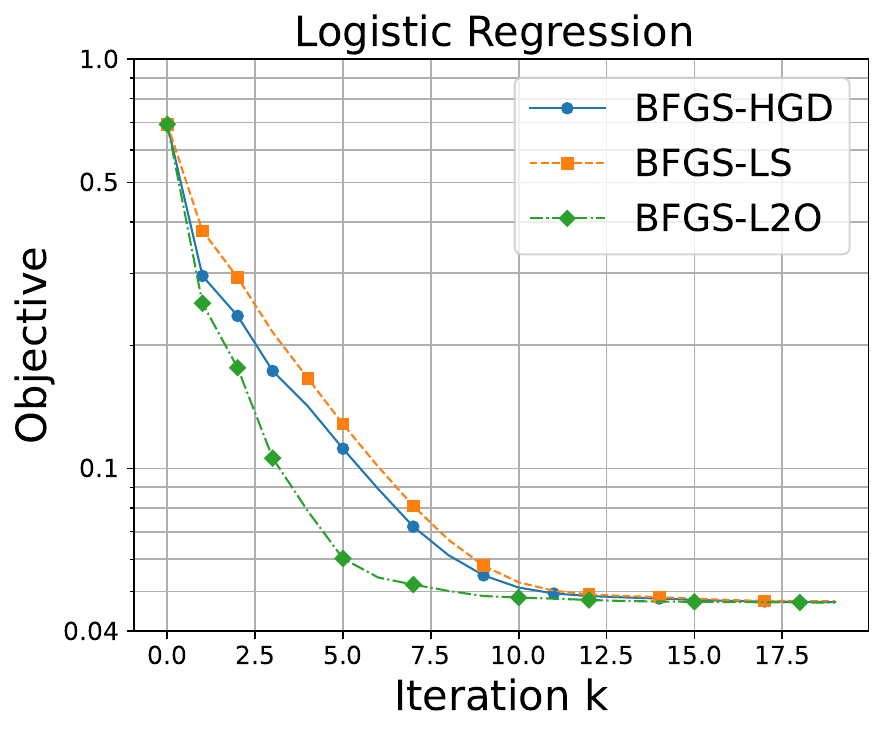}
        \vspace{-6mm}
        \caption{Logistic Regression. }
        \label{fig:logistic_regression}
    \end{minipage}%
    \hfill
    \begin{minipage}{0.32\textwidth}
        \centering
        \includegraphics[width=\textwidth]{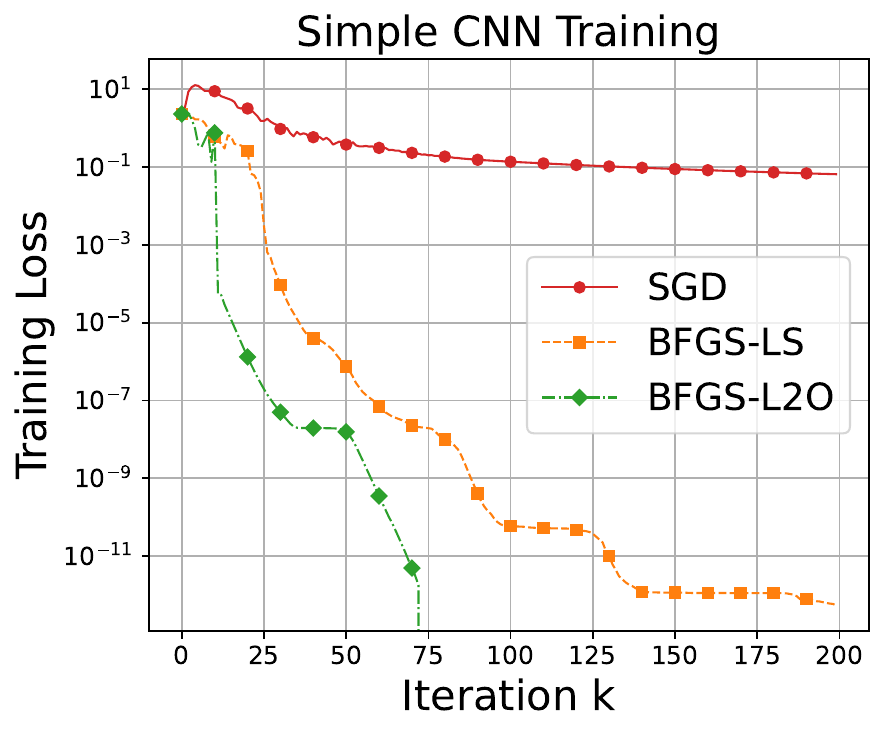}
        \vspace{-6mm}
        \caption{Simple CNN.}
        \label{fig:TinyCNN}
    \end{minipage}%
    \vspace{-3mm}
\end{figure*}

\begin{figure*}[tb]
    \centering
    \includegraphics[width=\linewidth]{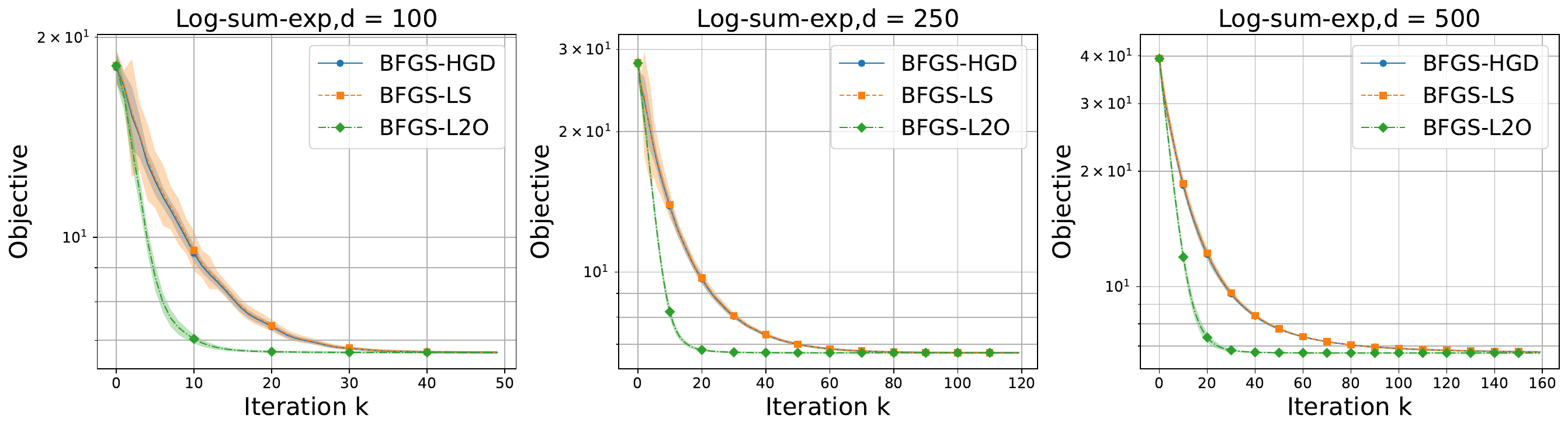}
    \vspace{-6mm}
    \caption{Log-sum-exp functions with different dimensions.}
    \label{fig:logsumexp}
    \vspace{-7mm}
\end{figure*}

% Our experiments were conducted using Python 3.9 and PyTorch 1.12 on an Ubuntu 18.04 system, equipped with an Intel Xeon Gold 5320 CPU and two NVIDIA RTX 3090 GPUs. 
We employed the Adam optimizer as our meta-optimizer to train our L2O model. For classic optimization problems, the training dataset consisted of 32,000 optimization problems with randomly sampled parameters, while a separate test dataset of 1,024 optimization problems was used for evaluation. Our L2O method (BFGS-L2O) was benchmarked against two baselines: Backtracking line search (BFGS-LS) and hypergradient descent (BFGS-HGD). 
% For BFGS-LS, the step size was initialized to 1 and iteratively scaled by a factor until the Armijo condition was satisfied. For BFGS-HGD, coordinate-wise step sizes were refined by performing 20 iterations of hypergradient descent within each BFGS iteration. 
All methods were tuned by experimenting with various parameter settings, and the best-performing configurations were selected for comparison. More details are provided in Appendix~\ref{appendix:exp}.

\paragraph{Least Squares Problems}

We first evaluate our L2O method on the classic least squares problems. The objective function is defined as:
    $\min_{x} f(x) = \frac{1}{2} \|Ax - b\|^2,$
where $A \in \mathbb{R}^{250\times 500}$ and $b \in \mathbb{R}^{500}$ are randomly generated using a Gaussian distribution. 
% Following the setting in \cite{liu2023towards}, we introduced sparsity by setting 90\% of the elements in matrix A to zero, resulting in a sparsity level of 0.1.

Figure \ref{fig:least_square} presents the convergence behavior for the least squares problems, where the optimization target is an objective value of zero. The optimization process was terminated when the gradient norm fell below $10^{-10}$. As depicted in the figure, BFGS-HGD offers a marginal improvement over BFGS-LS. In contrast, our proposed BFGS-L2O method demonstrates a significant reduction in convergence time, requiring approximately 60 iterations to reach the target, compared to around 80 iterations for BFGS-LS and BFGS-HGD, which represents an improvement of roughly 25\% enhancement in convergence speed. The nearly linear trajectory (on a log-scale for the objective value) of our BFGS-L2O method is consistent with superlinear convergence.

\paragraph{Logistic Regression Problems}

Next, we considered logistic regression problems for binary classification. The objective function is given by:
 $   \min_{x} f(x) = \frac{1}{m}\sum_{i=1}^{m} [b_i \log(h(a_i^Tx))
    +(1-b_i)\log(1-h(a_i^Tx))] + \rho \|x\|_2^2,$
where $m=500$, $\{(a_i,b_i)\in \mathbb{R}^{250} \times \{0,1\}\}_{i=1}^m$ are randomly generated, $h(z) = \frac{1}{1+e^{-z}}$ is the sigmoid function.

Figure \ref{fig:logistic_regression} illustrates the performance on logistic regression problems. While BFGS-HGD achieves a slightly lower objective function value than BFGS-LS during the initial iterations , both baseline methods exhibit similar overall convergence iteration counts, reaching the plateau around 15 iterations. In contrast, our proposed BFGS-L2O method shows notably faster convergence and consistently maintains a lower objective function value throughout the optimization process.

\paragraph{Log-Sum-Exp Problems}
For the log-sum-exp function, the objective function is:
$   \min_{x} f(x) = \log\left(\sum_{i=1}^{m} e^{a_i^Tx-b_i}\right),$
where $m=500$, $\{(a_i,b_i)\in \mathbb{R}^{d}\times \mathbb{R}\}_{i=1}^m$. 
% We followe the dataset generation process described in \cite{rodomanov2021greedy}. 
% We first generate auxiliary random vectors $\{ \hat{a_i}\}_{i=1}^m$ by sampling uniformly from the interval $[0,1]$. We then generate $\{b_i\}_{i=1}^m$ from the standard normal distribution. Using these, we define an auxiliary function $\hat{f}(x)= \log\left(\sum_{i=1}^{m} e^{\hat{a}_i^Tx-b_i}\right)$. Finally, we set $a_i = \hat{a}_i - \nabla \hat{f}(0)$, ensuring that the optimal solution of $f(x)$ is 0.

Figure \ref{fig:logsumexp} displays the results for log-sum-exp problems across different dimensions (d=100,250,500), revealing a clear trend: as the dimensionality increases, the performance advantage of our BFGS-L2O method over the baselines becomes more pronounced. For instance, when d=100, BFGS-L2O converges in approximately 20 iterations, whereas BFGS-LS and BFGS-HGD require around 40 iterations, resulting in a 2-fold speedup. This advantage grows with dimensionality; for d=250, BFGS-L2O converges in roughly 30 iterations compared to approximately 90-100 iterations for the baselines, achieving about a 3-fold speedup. The improvement is even more significant for d=500, where BFGS-L2O converges in about 40 iterations, while the baselines take around 150-160 iterations, demonstrating an approximate 4-fold enhancement in convergence speed. The figures also illustrate the variance (shaded areas) among different optimization runs. While the variance appears larger for all methods in lower dimensions (e.g., d=100), the variance of BFGS-L2O is consistently tighter compared to BFGS-LS across all tested dimensions, indicating greater stability for our proposed method.

\paragraph{Simple CNN Training}

To assess the performance of our method on a more complex optimization problem, we trained a simple CNN on the MNIST dataset. The network architecture comprised two convolutional layers followed by a fully connected layer. It is worth noting that while second-order methods like BFGS are highly effective for classical optimization problems, particularly those that are convex and smooth, they are less commonly employed for training deep neural networks due to computational costs and challenges with sophisticated landscapes. Our aim here is specifically to test the adaptability and robustness of BFGS-L2O under such demanding conditions. We found that the BFGS-HGD performed with high instability on this problem, and thus its performance is not included in the comparison. The training loss curves are presented in Figure \ref{fig:TinyCNN}. In this challenging scenario, our proposed BFGS-L2O method again demonstrates superior performance. It achieves a lower training loss more rapidly than both BFGS-LS and a standard SGD optimizer.

\section{Conclusions}
\label{sec:conclusion}
% This work investigates the use of coordinate-wise step sizes in BFGS method. We examine the advantages and challenges of identifying coordinate-wise step sizes by theoretical and numerical analysis. We rigorously derive conditions to enhance convergence properties. Building on these findings, we develop an L2O model that predicts optimal coordinate-wise step sizes. Experimental results demonstrate that our proposed approach outperforms baseline methods in terms of both convergence speed and stability.

This work investigated the application of coordinate-wise step sizes in the BFGS method. Through theoretical and numerical analyses, we examined the associated benefits and complexities. We rigorously derived sufficient conditions for coordinate-wise step size designed to enhance convergence properties. Building on this theoretical foundation, we developed a L2O model that effectively predicts these step sizes. Experimental results demonstrate that our proposed L2O approach significantly outperforms standard baseline methods in both convergence speed and stability.

{
    \small
    \bibliographystyle{plain}
    \bibliography{main}
}

% %%%%%%%%%%%%%%%%%%%%%%%%%%%%%%%%%%%%%%%%%%%%%%%%%%%%%%%%%%%%

\appendix
%!TEX root = neurips_2025.tex

\clearpage
\setcounter{page}{1}
% \maketitlesupplementary

\section{Proofs}
\label{appendix:proofs}
\subsection{Proof of Theorem \ref{theorem:bounds of smooth function}}
\label{proof:bounds of smooth function}
\begin{proof}
Consider a quadratic function 
\begin{equation*}
    Q(x) = f(y) + \nabla f(y)^\top (x-y) + \frac{\alpha}{2}(x-y)^\top B_k P_k^{-1}(x-y).
\end{equation*}

Basically, we use this quadratic function as a local approximation of $f(x)$ around $y$, and we think the minimum of this quadratic function is a better solution than $y$. This can work only if $Q(x)$ is an overestimation of $f(x)$. Indeed, we can show that:
\begin{align*}
&\quad Q(x) \\
&= f(y) + \nabla f(y)^\top (x-y) + \frac{\alpha}{2}(x-y)^\top B_k P_k^{-1}(x-y) \\
&\geq f(y) + \nabla f(y)^\top (x-y) + \frac{\alpha}{2} \frac{1}{\|B_k^{-1}\|_2\|P_k\|_2} \|x-y\|_2^2 \\
&\geq f(y) + \nabla f(y)^\top (x-y) + \frac{L}{2} \|x-y\|^2 \\
&\geq f(x),
\end{align*}
where the second inequality uses the condition $\|P_k\|_2\leq \frac{\alpha}{L\|B_k^{-1}\|_2}$ and the third uses the assumption of $L$-smoothness.

Plugging $x = y - P_k B_k^{-1} \nabla f(y)$ into the inequality, we get the Armijo condition:
\begin{equation*}
    f(y) - (1-\frac{\alpha}{2}) \nabla f(y)^\top P_k B_k^{-1} \nabla f(y) \geq f(y - P_k B_k^{-1} \nabla f(y)).
\end{equation*}

Let $x_k = y$ and $x_{k+1} = y - P_k B_k^{-1} \nabla f(y)$, we have
\begin{align*}
f(x_{k+1}) &\leq f(x_k) - (1-\frac{\alpha}{2}) \nabla f(x_k)^\top P_k B_k^{-1} \nabla f(x_k) \\
&\leq f(x_k) - (1-\frac{\alpha}{2}) \beta \frac{(\nabla f(x_k)^\top B_k^{-1} \nabla f(x_k))^2}{\|B_k \nabla f(x_k)\|_2^2} \\
&= f(x_k) - (\beta - \frac{\alpha\beta}{2}) \cos^2 \theta_k \|\nabla f(x_k)\|_2^2,
\end{align*}
where $\theta_k$ is the angle between $\nabla f(x_k)$ and $B_k^{-1} \nabla f(x_k)$. In the second inequality, we use the condition that $\lambda_{\max}(P_k) \geq \beta \frac{\nabla f(x_k)^TB_k^{-1}\nabla f(x_k)}{\|B_k^{-1}\nabla f(x_k)\|_2^2}$.

Following the proof of Theorem 3.2 in \cite{wright2006numerical}, summing over all iterations, we have:
\begin{equation*}
    f(x_{k+1}) \leq f(x_0) - (\beta - \frac{\alpha\beta}{2}) \sum_{i=0}^k \cos^2 \theta_i \|\nabla f(x_i)\|^2.
\end{equation*}

Note that $f(x_0)-f(x_{k+1})$ is lower-bounded by $0$ and upper-bounded by $f(x_0)$. Hence when $k$ approaches infinity, we have:
\begin{equation*}
    \sum_{i=0}^k \cos^2 \theta_i \|\nabla f(x_i)\|_2^2 < \infty,
\end{equation*}
which implies
\begin{equation*}
    \cos^2 \theta_k \|\nabla f(x_k)\|_2^2 \to 0.
\end{equation*}

Since $\|B_k\|_2\|B_k^{-1}\|_2 < M$, 
\begin{align*}
    \cos \theta_k &= \frac{\nabla f(x_k)^\top B_k^{-1} \nabla f(x_k)}{\|\nabla f(x_k)\|_2 \|B_k^{-1} \nabla f(x_k)\|_2} \\
    &\geq \frac{\lambda_{B_k^{-1}}^{\min} \|\nabla f(x_k)\|_2^2}{\lambda_{B_k^{-1}}^{\max} \|\nabla f(x_k)\|_2^2} \\
    & = \frac{1/\|B_k\|_2}{\|B_k^{-1}\|_2}\\
    &> \frac{1}{M}.
\end{align*}

Then we have
\begin{equation*}
    \lim_{k \to \infty} \|\nabla f(x_k)\|_2 = 0.
\end{equation*}

\end{proof}

\subsection{Proof of Theorem \ref{theorem:bounds of convex function}}
\label{proof:bounds of convex function}
\begin{proof}
Let $e_k = x_k -x^*$ denote the difference between the current iterate and the minimizer. We can write the update rule as:
\begin{equation*}
    e_{k+1} = e_k - P_k B_k^{-1} \nabla f(x_k).
\end{equation*}
Using the mean value theorem for vector-valued functions, We can write the gradient as:
\begin{align*}
    \nabla f(x_k) &= \nabla f(x_k) - \nabla f(x^*) \\
    &= \int_0^1 \nabla^2 f(x^* + te_k) e_k dt \\
    &= H_ke_k
\end{align*}
where:
\begin{equation*}
    H_k = \int_0^1 \nabla^2 f(x^* + te_k) dt.
\end{equation*}
Since $f$ is $L$-smooth , we have:
\begin{equation*}
     \nabla^2f(x) \preceq L I.
\end{equation*}
Note that $H_k$ is nothing else but the average of the Hessian matrix along the line segment between $x^*$ and $x_k$. Then we have:
\begin{align*}
     H_k \preceq L I.
\end{align*}
Substitute $\nabla f(x_k) = H_ke_k$ into the error update, we have:
\begin{align*}
    e_{k+1} &= e_k - P_k B_k^{-1} H_k e_k \\
    &= (I - P_k B_k^{-1} H_k) e_k.
\end{align*}
Consider the matrix $T_k = I - P_k B_k^{-1} H_k$, we will analyze the spectral radius of it. The upper bound of the eigenvalue of $T_k$ is 1, since $P_k$ is a diagonal matrix with positive entries $p_{k,i}$, and $B_k^{-1}$ and $H_k$ are positive definite matrices. The lower bound of the eigenvalue of $T_k$ is:
\begin{align*}
    \lambda_{min}(T_k) &= 1 - \lambda_{max}(P_k) \lambda_{max}(B_k^{-1}) \lambda_{max}(H_k)\\
    &\geq 1 - \frac{2 \gamma}{L} \frac{L}{\gamma}  = -1 
\end{align*}
Hence the spectral radius of $T_k$ is less than 1. Then we have:
\begin{equation*}
    \|e_{k+1}\|_2 \leq \|T_k\|_2\|e_k\|_2 \leq \|e_k\|_2.
\end{equation*}
\end{proof}

\subsection{Proof of Theorem \ref{theorem:limit of P}}
\label{proof:limit of P}
\begin{proof}
We first show that if 
\begin{equation}
\label{eq:appendix_1}
    \lim_{k \to \infty} \frac{\|[B_k P_k^{-1} - \nabla^2 f(x^*)](x_{k+1} - x_k)\|}{\|x_{k+1} - x_k\|} = 0,
\end{equation}

then $\{x_k\}$ converges to $x^*$ superlinearly.

We can write the numerator as

\[
\begin{aligned}
    & \relax [B_k P_k^{-1} - \nabla^2 f(x^*)](x_{k+1} - x_k) \\
    &= - \nabla f(x_k) - \nabla^2 f(x^*)(x_{k+1} - x_k) \\
    &= \nabla f(x_{k+1}) - \nabla f(x_k) \\
    &\quad - \nabla^2 f(x^*)(x_{k+1} - x_k) - \nabla f(x_{k+1}) \\
    &= \nabla^2 f((1 - \xi) x_{k+1} + \xi x_k) (x_{k+1} - x_k) \\
    &\quad - \nabla^2 f(x^*)(x_{k+1} - x_k) - \nabla f(x_{k+1}),
\end{aligned}
\]

where the last equation uses the mean value theorem and $\xi \in (0,1)$.

Since $\{x_k\}$ converges to $x^*$, we have:
\begin{align*}
    &\lim_{k \to \infty} \frac{\nabla^2 f((1 - \xi) x_{k+1} + \xi x_k) (x_{k+1} - x_k) }{x_{k+1} - x_k} \\ 
    &\quad - \frac{- \nabla^2 f(x^*)(x_{k+1} - x_k)}{x_{k+1} - x_k} \\
    &= \lim_{k \to \infty} \nabla^2 f((1 - \xi) x_{k+1} + \xi x_k) - \nabla^2 f(x^*) = 0.
\end{align*}

Hence,
\begin{equation*}
    \lim_{k \to \infty} \frac{\|\nabla f(x_{k+1})\|}{\|x_{k+1} - x_k\|} = 0.
\end{equation*}

Since $\nabla^2 f(x^*)$ is nonsingular, there exists a $\beta > 0$ such that
\begin{align*}
    \|\nabla f(x_{k+1})\| &= \|\nabla f(x_{k+1}) - \nabla f(x^*)\| \\
    &= \|\nabla^2 f(x_{\xi}) (x_{k+1} - x^*)\| \\
    &\geq \beta \|x_{k+1} - x^*\|.
\end{align*}

Therefore,
\begin{align*}
    \frac{\nabla f(x_{k+1})}{\|x_{k+1} - x_k\|} &\geq \frac{\beta \|x_{k+1} - x^*\|}{\|x_{k+1} - x_k\|} \\
    &\geq \frac{\beta \|x_{k+1} - x^*\|}{\|x_{k+1} - x^*\| + \|x_k - x^*\|} \\
    &= \beta \frac{\frac{\|x_{k+1} - x^*\|}{\|x_k - x^*\|}}{1 + \frac{\|x_{k+1} - x^*\|}{\|x_k - x^*\|}}.
\end{align*}

Hence, $\lim_{k \to \infty} \frac{\|x_{k+1} - x^*\|}{\|x_k - x^*\|} = 0$, which indicates superlinear convergence. 

Then we show that \eqref{eq:appendix_1} is indeed true. Theorem 3.4 in \cite{dennis1974characterization} shows that the $\{B_k\}$ generated by BFGS satisfies
\begin{equation*}
    \lim_{k \to \infty} \frac{\|B_k - \nabla^2 f(x^*)\|_2}{\|x_{k+1} - x_k\|} = 0.
\end{equation*}
If $P_k$ converges to the identity matrix,
\begin{align*}
    &\quad~\lim_{k \to \infty} \frac{\|[B_k P_k^{-1} - \nabla^2 f(x^*)](x_{k+1} - x_k)\|}{\|x_{k+1} - x_k\|} \\
    &=\lim_{k \to \infty} \frac{\|[B_k  - \nabla^2 f(x^*)](x_{k+1} - x_k)\|}{\|x_{k+1} - x_k\|} \\
    &\leq \lim_{k \to \infty} \frac{\|B_k  - \nabla^2 f(x^*)\|_2\|(x_{k+1} - x_k)\|}{\|x_{k+1} - x_k\|}\\
    & = 0.
\end{align*}
And this limitation is obviously non-negative, hence it is zero. Then we can conclude that $\{x_k\}$ converges to $x^*$ superlinearly. 
\end{proof}

\section{Detailed Experimental Setup}
\label{appendix:exp}

This section provides supplementary details regarding our experimental settings, including the computational environment, dataset generation procedures, L2O model training, baseline configurations, and the specifics of the neural network training task.

\subsection{Computational Environment}
Our experiments were conducted using Python 3.9 and PyTorch 1.12. The underlying system was Ubuntu 18.04, equipped with an Intel Xeon Gold 5320 CPU and two NVIDIA RTX 3090 GPUs.

\subsection{L2O Model Training (BFGS-L2O)}
The L2O parameters for our BFGS-L2O model were trained using Adam as the meta-optimizer. The Adam learning rate was set to $1\times 10^{-3}$, and we processed a batch size of 64 optimization problems for each update. The L2O model underwent a total of 200 such training updates.

\subsection{Datasets for Classic Optimization Problems}
For the classic optimization problems (Least Squares and Log-Sum-Exp), the L2O training dataset consisted of 32,000 randomly generated problem instances. A separate test dataset of 1,024 instances was used for evaluation. Specific parameter generation for each problem type is detailed below. 

\paragraph{Least Squares Problem}
The objective function is:
\begin{equation*}
\min_{x} f(x) = \frac{1}{2} |Ax - b|^2,
\end{equation*}
where $A\in \mathbb R^{m\times n}$ and $b\in \mathbb R^n$. For our experiments, we used m=250 and n=500. The elements of A and b were randomly generated using a Gaussian distribution. Following the setup in \cite{liu2023towards} (from your main text), sparsity was introduced into A by setting 90\% of its elements to zero.

\paragraph{Log-Sum-Exp Problem}
The objective function is:
\begin{equation*}
\min_{x} f(x) = \log\left(\sum_{i=1}^{m} e^{a_i^Tx-b_i}\right) ,
\end{equation*}
where $m=500$ (number of exponential terms). The vectors $\{(a_i,b_i)\in \mathbb R^d \times \mathbb R\}_{i=1}^m $ were generated following the dataset generation process described in \cite{rodomanov2021greedy} to ensure the optimal solution is $x^*=0$. We first generate auxiliary random vectors $\{ \hat{a_i}\}_{i=1}^m$ by sampling uniformly from the interval $[0,1]$. We then generate $\{b_i\}_{i=1}^m$ from the standard normal distribution. Using these, we define an auxiliary function $\hat{f}(x)= \log\left(\sum_{i=1}^{m} e^{\hat{a}_i^Tx-b_i}\right)$. Finally, we set $a_i = \hat{a}_i - \nabla \hat{f}(0)$, ensuring that the optimal solution of $f(x)$ is 0.

\subsection{Baseline Method Configurations}
\begin{itemize}
\item BFGS-LS (BFGS with Backtracking Line Search): The step size was initialized to 1 at each iteration. The backtracking line search iteratively scaled the step size by 0.8 until the Armijo condition $f(x_k+\alpha_k d_k)\leq f(x_k)+c_1 \alpha_k \nabla f(x_k)^Td_k$ is satisfied. Here $d_k$ is the descent direction and $c_1=10^{-4}$. 
\item BFGS-HGD (BFGS with Hypergradient Descent): Within each BFGS iteration, coordinate-wise step sizes $P_k$ were initialized as the identity matrix and then refined by performing 20 iterations of hypergradient descent with hyper step size $\eta=10^{-2}$.
\end{itemize}

\subsection{Simple CNN Training Details}
The Convolutional Neural Network (CNN) used for the MNIST dataset experiments processes input images of size $28\times 28\times 1$. The architecture begins with a first convolutional layer applying 2 filters with a $3\times 3$ kernel, stride 1, and padding 1, followed by a ReLU activation, resulting in a $28\times 28\times 2$ volume. This is then downsampled by a $2\times 2$ max pooling layer with a stride of 2, producing a $14\times 14\times 2$ volume. A second convolutional layer follows, applying 3 filters with a $3\times 3$ kernel, stride 1, and padding 1, again followed by ReLU activation, yielding a $14\times 14\times 3$ volume. This is further downsampled by a second $2\times 2$ max pooling layer with a stride of 2, resulting in a $7\times 7\times 3$ volume. This output is then flattened into a vector of 147 features, which feeds into a fully connected layer that produces 10 output units, corresponding to the logits for the 10 MNIST classes.

\section{Gain of CWSS}
\label{appendix:gain_of_cwss}

To illustrate the potential benefits of \cwss in the BFGS method, let us consider the theoretical implications of relaxing the constraint that step size should be a scalar. Assume we have identified an optimal scalar step size, denoted by $\alpha_k^*$, for the $k$-th iteration. 
Since the restriction of a convex function to a line remains convex \cite{boyd2004convex}, this optimal step size $\alpha_k^*$ satisfies:
\begin{equation}
\label{eq:scalar_step_size}
\begin{aligned}
&\frac{d}{d \alpha_k} f(x_{k+1}) \bigg|_{\alpha_k = \alpha_k^*} \\
=& \frac{d}{d\alpha_k^*} f(x_k - \alpha_k ^*B_k^{-1} \nabla f(x_k)) \\
=& -\nabla f(x_k - \alpha_k^* B_k^{-1} \nabla f(x_k))^\top B_k^{-1} \nabla f(x_k) \\
=& 0.
\end{aligned}
\end{equation}
When constrained to a scalar form, $\alpha_k^*$ guarantees optimality along the single search direction $B_k^{-1} \nabla f(x_k)$. However, if we allow the step size to be a diagonal matrix $P_k$ rather than a scalar, the optimality condition of $\alpha_k^*$ may no longer hold. 
By extending to a coordinate-wise approach, we aim to further minimize the objective function by adjusting each coordinate independently, which can potentially achieve a lower function value than with $\alpha_k^*$ alone.
To explore this, let $P_k = \alpha_k^* I$, and consider the partial derivative of $f(x_{k+1})$ with respect to $P_k$ at this point:
\begin{equation}
\label{eq:partial_derivative}
\begin{aligned}
\frac{\partial}{\partial P_k} f(x_k - P_k B_k^{-1} \nabla f(x_k)) \bigg|_{P_k = \alpha_k^* I} &= \\
\text{diag}( -\nabla f(x_k - \alpha_k^* B_k^{-1} \nabla f(x_k)) \odot &B_k^{-1} \nabla f(x_k)),
\end{aligned}
\end{equation}
where $\odot$ denotes the Hadamard (element-wise) product. Since $B_k^{-1}\nabla f(x_k)\neq 0$ , the derivative in \eqref{eq:partial_derivative} equals zero only if $\nabla f(x_k - \alpha_k^* B_k^{-1} \nabla f(x_k)) = 0$. Since the optimum does not generally lie on the direction of $B_k^{-1}\nabla f(x_k)$, the dot product being zero in \eqref{eq:scalar_step_size} does not imply that the Hadamard product is also zero in \eqref{eq:partial_derivative}. This observation suggests that even we know the optimal scalar step size $\alpha^*$, we can still find coordinate-wise step sizes that could achieve a more effective descent.
To determine suitable coordinate-wise step sizes, we employ hypergradient descent. Defining $g(p) = f(x_k - p \odot B_k^{-1} \nabla f(x_k))$, where $p$ is the diagonal of $P$,we can analyze the smoothness of $g(p)$ as follows:
\begin{align*}
&\quad \|\nabla g(p_1) - \nabla g(p_2)\| \\
&= \|(\nabla f(x_k - p_1 \odot B_k^{-1} \nabla f(x_k)) \\
&\quad\quad - \nabla f(x_k - p_2 \odot B_k^{-1} \nabla f(x_k))\| \\
&\leq L \|(p_1 - p_2) \odot B_k^{-1} \nabla f(x_k)\| \\
&\leq L \|B_k^{-1} \nabla f(x_k)\| \|p_1 - p_2\| \\
&\leq LR \|p_1 - p_2\|,
\end{align*}
where $L$ is the Lipschitz constant of $\nabla f$ and $R$ is from assumption~\ref{assumption:bounded_update_direction}. This shows that $g(p)$ is $LR$-smooth. 
To explore this, we can set the coordinate-wise step sizes $P_k$ as:
\begin{equation*}
P_k = \alpha_k^* I - \frac{1}{LR} v_k B_k^{-1} \nabla f(x_k),
\end{equation*}
where $v_k = \text{diag}(\nabla f(x_k - \alpha_k^* B_k^{-1} \nabla f(x_k)))$, $L$ is the Lipschitz constant of $\nabla f$ and $R$ is from assumption~\ref{assumption:bounded_update_direction}. This coordinate-wise step size $P_k$ is theoretically guaranteed to perform better than the scalar step size $\alpha_k^*$:
\begin{equation*}
\begin{split}
&f(x_k-P_kB_k^{-1}\nabla f(x_k)) \leq f(x_k-\alpha_k^*B_k^{-1}\nabla f(x_k)) \\
&\quad - \frac{1}{2LR} |\nabla f(x_k-\alpha_k^*B_k^{-1}\nabla f(x_k))\odot B_k^{-1}\nabla f(x_k) |^2.
\end{split}
\end{equation*}

This demonstrates that \cwss in the BFGS method can yield a more substantial decrease in the objective function than a scalar step size. 

\section{Limitations}
\label{appendix:limitations}
Our theoretical framework provides a solid foundation for the design of \cwss, but its derivations rely on assumptions such as L-smoothness and specific properties of the Hessian approximations. These assumptions may not universally hold in all practical optimization scenarios, potentially impacting the direct applicability of the derived theoretical guarantees. Furthermore, while our theory provides bounds for \cwss, their practical implementation involves a simplification, constraining step sizes to the interval from 0 to 2, because accurately estimating problem-dependent quantities like Lipschitz constants or Hessian conditioning in real-time during optimization is often infeasible. Consequently, although our L2O model learns an effective step-size strategy within this theoretically defined "safe operating region" to ensure stability, this constrained approach might theoretically preclude the discovery of a globally optimal or even faster-converging step-size strategy if such a strategy were to exist outside these established bounds. Despite these theoretical considerations, our proposed L2O method demonstrates significant performance improvements and practical value across a range of experiments, underscoring its potential in tackling complex optimization problems.

\section{Impact Statement}
\label{appendix:impact_statment}
This work concentrates on advancing optimization algorithms by leveraging deep learning techniques, aiming to enhance their computational efficiency and robustness. Potential positive societal impacts stem from enabling more effective and resource-conscious solutions to complex computational problems across diverse scientific and engineering fields that rely on optimization. While this research is foundational to optimization methodology, it is important to acknowledge that, like any significant improvement in computational tools, the resulting enhanced capabilities should be applied responsibly.

% %%%%%%%%%%%%%%%%%%%%%%%%%%%%%%%%%%%%%%%%%%%%%%%%%%%%%%%%%%%%

% \newpage
% \input{sec/neurips_paper_checklist}

\end{document}